\documentclass[12pt,a4paper]{article}

\usepackage{cite}
\usepackage[latin1]{inputenc}
\usepackage{graphicx,array,multirow}
\usepackage{amsmath}
\usepackage{amsfonts}
\usepackage{amssymb}
\usepackage{makeidx}
\usepackage{setspace}
\usepackage{geometry}
\usepackage{color}
\usepackage{hyperref}
\usepackage[table]{xcolor}
\usepackage{natbib}
\usepackage{wrapfig}
\usepackage{epigraph}
\hypersetup{colorlinks, linkcolor={red!50!black}, citecolor={blue!50!black}, urlcolor={blue!80!black}}
\title{Compositional generalization in semantic parsing with pretrained transformers}
\author{%
  Emin Orhan \\
  New York University\\
  \texttt{eo41@nyu.edu}
  }
\date{}

\begin{document}

\maketitle

\begin{abstract}
Large-scale pretraining instills large amounts of knowledge in deep neural networks. This, in turn, improves the generalization behavior of these models in downstream tasks. What exactly are the limits to the generalization benefits of large-scale pretraining? Here, we report observations from some simple experiments aimed at addressing this question in the context of two semantic parsing tasks involving natural language, SCAN and COGS. We show that language models pretrained exclusively with non-English corpora, or even with programming language corpora, significantly improve out-of-distribution generalization in these benchmarks, compared with models trained from scratch, even though both benchmarks are English-based. This demonstrates the surprisingly broad transferability of pretrained representations and knowledge. Pretraining with a large-scale protein sequence prediction task, on the other hand, mostly deteriorates the generalization performance in SCAN and COGS, suggesting that pretrained representations do not transfer universally and that there are constraints on the similarity between the pretraining and downstream domains for successful transfer. Finally, we show that larger models are harder to train from scratch and their generalization accuracy is lower when trained up to convergence on the relatively small SCAN and COGS datasets, but the benefits of large-scale pretraining become much clearer with larger models.
\end{abstract}

\section{Introduction}
\epigraph{\textit{here is the root of the root and the bud of the bud
and the sky of the sky of a tree called life}}{--- e e cummings}

Large, pretrained language and vision models are widely used in a diverse range of downstream NLP and computer vision tasks. These models encode a large amount of transferable knowledge (of varying specificity) in their parameters and display remarkable (sometimes even surprising) ``emergent'' generalization behaviors as a result. To give a few examples: (i) the highly influential GPT-3 model displays hallmarks of few-shot learning/generalization ability from a small number of examples given in its prompt \citep{brown2020}; (ii) DALL-E, or other similar text-to-image models trained on large datasets such as CLIP+VQGAN, display qualitative evidence of compositional generalization abilities \citep{ramesh2021,crowson2022}; (iii) pretraining with large, diverse image or text datasets improves the out-of-distribution generalization performance of image recognition \citep{orhan2019,xie2020} and NLP models \citep{hendrycks2020}, respectively.

What exactly are the limits to the generalization benefits of large-scale pretraining? More concretely, how do pretraining benefits scale with factors such as the diversity and the scale of the pretraining data, the model size, or the similarity between the pretraining data and the downstream task? Here, we report the results of some simple experiments aimed at addressing the latter two factors, namely the model size and the similarity between the pretraining data and the downstream task, in the context of two previously introduced semantic parsing tasks, SCAN \citep{lake2018} and COGS \citep{kim2020}.  

A few other recent works have also attempted to address some of these questions. \cite{furrer2020} and \cite{tay2021} show that large-scale pretraining ---in the form of language modeling on the web-scale C4 corpus \citep{raffel2020} consisting mainly of English language texts--- can improve compositional generalization in SCAN and COGS benchmarks, respectively. With respect to the specificity of pretraining benefits, \cite{lu2021} argue that pretraining with language modeling confers broad benefits in downstream tasks, including, somewhat surprisingly, a variety of non-language tasks such as image recognition or protein sequence modeling. \cite{papadimitriou2020} similarly report downstream natural language modeling tasks can benefit from pretraining in seemingly unrelated domains such Java code or MIDI music scores. They argue that abstract syntactic similarity between the pretraining and downstream domains is key for the transfer success of pretraining. Consistent with this idea, \cite{chiang2020} also show that pretraining with a simple artificial language generated by a stack-based hierarchical grammar can improve the accuracy on a diverse set of downstream natural language tasks, namely the tasks comprising the GLUE benchmark \citep{wang2018}. More recently, \cite{krishna2021} show that text summarization also does not seem to require pretraining with natural language texts: they demonstrate that even pretraining with texts consisting entirely of randomly and independently sampled nonsense words achieves similar results in downstream text summarization tasks. In the visual domain, \cite{baradad2021} show that even very simple random noise processes can be used as effective pretraining data for downstream natural image recognition tasks, as long as these noise processes satisfy certain basic structural properties of natural images. In a similar vein, \cite{sinha2021} recently showed that scrambling the (within-sentence) word order in natural language texts has surprisingly little effect on their effectiveness as pretraining data, as long as some higher-order co-occurrence statistics are preserved.

The results we report here contribute valuable observations to this prior literature. Our main results can be summarized as follows: 

\begin{itemize}
\item Large-scale pretraining with natural language based text-to-text tasks improve compositional generalization in both SCAN and COGS benchmarks. This result is essentially a replication of earlier reports to the same effect: e.g. \cite{furrer2020} for SCAN and \cite{tay2021} for COGS.

\item Surprisingly, however, even models pretrained exclusively with non-English languages significantly improve performance on these benchmarks compared to models trained from scratch, even though both benchmarks are English-based. 

\item Even more surprisingly, a language model pretrained predominantly on programming languages \citep{wang2021} also provides large generalization benefits on both SCAN and COGS, roughly equivalent in size to the generalization benefits of large-scale pretraining with natural languages.

\item However, the same is not true for pretraining with a large-scale protein sequence prediction task \citep{elnaggar2020}. Pretraining with this task in fact mostly deteriorates the performance on the downstream semantic parsing tasks, suggesting that pretrained representations do not transfer universally and that there are likely constraints on the similarity between the pretraining and downstream tasks for successful transfer.

\item Bigger models are harder to train from scratch (and their generalization accuracy lower when trained to convergence) on the relatively small-scale SCAN and COGS benchmarks, but they benefit more from large-scale pretraining.
\end{itemize}

\section{Tasks}
We evaluated the potential benefits of large-scale pretraining for (out-of-distribution) compositional generalization in two semantic parsing tasks, SCAN and COGS. We now briefly describe these benchmarks. We refer the reader to the original papers for further details.

\textbf{SCAN \citep{lake2018}}: In SCAN, the goal is to map a procedurally generated instruction in simplified natural language (English) to a more formal semantic representation in a domain-specific language: e.g. \textit{turn left twice} $\rightarrow$ LTURN LTURN. The benchmark consists of several different train/test splits evaluating different aspects of compositional generalization. In this work, we considered all SCAN splits evaluated in \cite{furrer2020} (seven different splits in total; see Table~\ref{scan_table}).

\textbf{COGS \citep{kim2020}}: COGS is similar to SCAN at a high level: like SCAN, it involves mapping procedurally generated English sentences to an abstract logical form. Unlike SCAN, however, it is geared more toward investigating compositional phenomena in natural language, hence the surface form English sentences in COGS are more complex and realistic and the targeted compositional phenomena are more linguistic in nature. COGS has a single training set and an out-of-distribution generalization set containing several different types of examples for evaluating different compositional phenomena in natural language (in English, more specifically). 

We have found that for both SCAN and COGS, training the models up to convergence on the training set always improved the average out-of-distribution generalization accuracy (over all splits/conditions), as has been observed recently by \cite{csordas2021} for models trained from scratch (in SCAN, some splits benefited from early stopping, but the average accuracy over all splits always favored training the model up to convergence). We therefore followed this strategy for all experiments reported below. This means that no early stopping, no cross-validation over a validation set, and no hyperparameter searches were performed in any of the experiments. All experiments used the example \texttt{translation} code provided in the Huggingface Examples repository\footnote{\url{https://github.com/huggingface/transformers/tree/master/examples/pytorch/translation}} with minimal modifications. All code for reproducing the experiments reported here, as well as the corresponding output files can be found at: \url{https://github.com/eminorhan/parsing-transformers}.

\section{Results}

\begin{table}
\centering
\begin{tabular}{cccccccccc}
 & \texttt{model} & \texttt{aj} & \texttt{atl} & \texttt{jar} & \texttt{ar} & \texttt{or} & \texttt{right} & \texttt{length} & \texttt{ave.} \\
\hline
\rowcolor{lightgray} \multirow{2}{*}{\rotatebox[origin=c]{90}{\texttt{mar.}}} & \texttt{marian\_defr\_scr} & 2.9 {\tiny$\pm$ 0.5} & 95.2 {\tiny$\pm$ 1.0} & 100.0 {\tiny$\pm$ 0.0}& 33.1 {\tiny$\pm$ 0.5} & 89.8 {\tiny$\pm$ 2.6} & 82.1 {\tiny$\pm$ 4.1} & 12.8 {\tiny$\pm$ 0.2} & 59.4   \\
& \texttt{marian\_defr} & 62.7 {\tiny$\pm$ 2.7} & 83.5 {\tiny$\pm$ 2.9} & 96.3 {\tiny$\pm$ 0.6} & 45.0 {\tiny$\pm$ 1.9} & 99.1 {\tiny$\pm$ 0.1} & 92.4 {\tiny$\pm$ 0.8} & 15.0 {\tiny$\pm$ 0.1} & 70.6  \\
\hline
\rowcolor{lightgray} \multirow{5}{*}{\rotatebox[origin=c]{90}{\texttt{t5-base}}} & \texttt{t5\_base\_scr} & 0.0 {\tiny$\pm$ 0.0} & 67.6 {\tiny$\pm$ 11.0} & 100.0 {\tiny$\pm$ 0.0} & 18.1 {\tiny$\pm$ 3.4} & 51.1 {\tiny$\pm$ 7.3} & 58.7 {\tiny$\pm$ 5.7} & 7.5 {\tiny$\pm$ 1.3} & 43.3   \\
& \texttt{t5\_base} & 93.6 {\tiny$\pm$ 0.6} & 55.2 {\tiny$\pm$ 2.3} & 93.6 {\tiny$\pm$ 0.0} & 38.4 {\tiny$\pm$ 0.9} & 96.8 {\tiny$\pm$ 0.3} & 65.3 {\tiny$\pm$ 3.0} & 13.5 {\tiny$\pm$ 0.1} & 65.2  \\
& \texttt{mt5\_base} & 9.7 {\tiny$\pm$ 4.4} & 96.4 {\tiny$\pm$ 1.4} & 99.8 {\tiny$\pm$ 0.1} & 16.7 {\tiny$\pm$ 5.2} & 85.9 {\tiny$\pm$ 6.7} & 73.3 {\tiny$\pm$ 5.5} & 10.7 {\tiny$\pm$ 0.6} & 56.1  \\
& \texttt{ct5\_base} & 5.2 {\tiny$\pm$ 0.4} & 98.2 {\tiny$\pm$ 1.6} & 100.0 {\tiny$\pm$ 0.0} & 62.1 {\tiny$\pm$ 12.6} & 94.2 {\tiny$\pm$ 4.7} & 100.0 {\tiny$\pm$ 0.0} & 5.0 {\tiny$\pm$ 0.5} & 66.4  \\
& \texttt{pt5\_base} & 0.0 {\tiny$\pm$ 0.0} & 79.2 {\tiny$\pm$ 4.2} & 95.7 {\tiny$\pm$ 2.0} & 2.4 {\tiny$\pm$ 0.9} & 23.4 {\tiny$\pm$ 8.5} & 15.1 {\tiny$\pm$ 3.4} & 11.6 {\tiny$\pm$ 0.4} & 32.5 \\
\hline
\rowcolor{lightgray} \multirow{4}{*}{\rotatebox[origin=c]{90}{\texttt{t5-3b}}} &\texttt{t5\_3b\_scr} & 0.0 {\tiny$\pm$ 0.0} & 8.4 {\tiny$\pm$ 0.4} & 11.7 {\tiny$\pm$ 0.4} & 0.0 {\tiny$\pm$ 0.0} & 0.5 {\tiny$\pm$ 0.4} & 0.2 {\tiny$\pm$ 0.1} & 0.6 {\tiny$\pm$ 0.1} & 3.3  \\
& \texttt{t5\_3b} & 96.1 {\tiny$\pm$ 1.4} & 94.5 {\tiny$\pm$ 4.4} & 100.0 {\tiny$\pm$ 0.0} & 37.0 {\tiny$\pm$ 1.0} & 99.5 {\tiny$\pm$ 0.1} & 98.6 {\tiny$\pm$ 0.4} & 8.3 {\tiny$\pm$ 0.7} & 76.3  \\
& \texttt{mt5\_xl} & 40.0 {\tiny$\pm$ 8.1} & 90.0 {\tiny$\pm$ 4.6} & 99.9 {\tiny$\pm$ 0.0} & 90.0 {\tiny$\pm$ 3.3} & 98.5 {\tiny$\pm$ 1.1} & 99.4 {\tiny$\pm$ 0.3} & 5.1 {\tiny$\pm$ 0.2} & 74.7  \\
& \texttt{pt5\_xl} & 0.1 {\tiny$\pm$ 0.1} & 50.9 {\tiny$\pm$ 5.2} & 84.0 {\tiny$\pm$ 13.1} & 4.7 {\tiny$\pm$ 3.0} & 42.3 {\tiny$\pm$ 10.2} & 9.0 {\tiny$\pm$ 2.8} & 11.3 {\tiny$\pm$ 1.8} & 28.9 \\
\hline
\end{tabular}
\caption{Exact match accuracies in different SCAN splits. Models with a \texttt{scr} attached to their name (highlighted in gray) are models trained from scratch (no pretraining), the other models are all pretrained with different datasets/tasks. Abbreviations of SCAN splits: \texttt{aj}: add jump, \texttt{atl}: add turn left, \texttt{jar}: jump around right, \texttt{ar}: around right, \texttt{or}: opposite right. The top two rows show the results for the Marian neural machine translation models (German-to-French); these models have $\sim$74M parameters. The middle four rows show the results for the \texttt{t5\_base} sized models; these models have $\sim$220M parameters. The bottom four rows show the results for the \texttt{t5\_3b} (or \texttt{xl}) sized models; these models have $\sim$3B parameters.~Note that the multilingual \texttt{mt5} models use different tokenizers from the corresponding \texttt{t5} models, hence their parameter counts are slightly different. Numbers in small font indicate the standard errors over at least three independent runs for each condition. The last column shows the average of averages over all splits.}
\label{scan_table}
\end{table}

\begin{table}
\centering
\begin{tabular}{ccc}
 & \texttt{model} & \texttt{average} \\ 
\hline 
\rowcolor{lightgray} \multirow{2}{*}{\rotatebox[origin=c]{90}{\texttt{mar.}}} & \texttt{marian\_defr\_scr} & 62.7 {\tiny$\pm$ 0.5} \\
 & \texttt{marian\_defr} & 83.4 {\tiny$\pm$ 0.1} \\ 
\hline 
\rowcolor{lightgray} \multirow{5}{*}{\rotatebox[origin=c]{90}{\texttt{t5-base}}} & \texttt{t5\_base\_scr} &  32.3 {\tiny$\pm$ 2.2} \\ 
 & \texttt{t5\_base} & 83.3 {\tiny$\pm$ 0.1}  \\
 & \texttt{mt5\_base} & 83.4 {\tiny$\pm$ 0.1}  \\
 & \texttt{ct5\_base} & 82.6 {\tiny$\pm$ 0.1}  \\
 & \texttt{pt5\_base} & 16.1 {\tiny$\pm$ 2.3} \\ 
\hline 
\rowcolor{lightgray} \multirow{4}{*}{\rotatebox[origin=c]{90}{\texttt{t5-3b}}} & \texttt{t5\_3b\_scr} & 15.5 {\tiny$\pm$ 0.6} \\
 & \texttt{t5\_3b} & 84.1 {\tiny$\pm$ 0.2} \\
 & \texttt{mt5\_xl} & 84.6 {\tiny$\pm$ 0.1} \\
 & \texttt{pt5\_xl} & 0.0 {\tiny$\pm$ 0.0} \\ 
\hline
\end{tabular}
\caption{Exact match accuracies on the COGS generalization set.~Numbers in small font are standard errors over at least three independent runs for each model.~Breakdown of accuracy into different conditions can be found in the output files provided on the accompanying github repository.}
\label{cogs_table}
\end{table}

{\bf 1. Pretraining improves compositional generalization:} Pretraining in natural language based tasks consistently improved the out-of-distribution generalization accuracy in both SCAN and COGS (Tables~\ref{scan_table}-\ref{cogs_table}; breakdown of accuracy into different conditions in COGS can be found in the output files provided on the accompanying github repository). For example, the pretrained T5-base (\texttt{t5\_base}) model \citep{raffel2020} outperformed the same model architecture trained from scratch (\texttt{t5\_base\_scr}) by $\sim$22\% on SCAN (averaged over all splits) and by $\sim$51\% on COGS in exact match accuracy (in absolute terms). These results reproduce earlier observations of similar pretraining benefits for (out-of-distribution) compositional generalization in SCAN and COGS \citep{furrer2020,tay2021}.

{\bf 2. Pretraining benefits are not language specific:} \cite{furrer2020} suggest that the main benefit of pretraining might be to improve the model's ability to substitute similar words or phrases with each other. This hypothesis is circumstantially supported by the observation that in SCAN, for instance, pretraining benefits are largest in splits requiring a single-word substitution (e.g. \texttt{add jump}) and smaller in splits requiring the substitution of longer phrases. However, this hypothesis assumes that pretraining with English language texts should be crucial for transfer success in SCAN and COGS. The pretrained T5 models from \cite{raffel2020} were indeed trained primarily on the C4 corpus, which consists mainly of English language texts. To see if pretraining benefits would be sustained with models trained primarily on languages other than English, we conducted the same experiments with the multilingual T5 (mT5) models \citep{xue2021}, which are language models trained on 101 different languages. Importantly, \cite{xue2021} report that less than 5\% of the training data for the mT5 models were English sentences. Despite this, we find that pretrained mT5 models (\texttt{mt5\_base} and \texttt{mt5\_xl}) also significantly improve generalization accuracy in both SCAN and COGS (Tables~\ref{scan_table}-\ref{cogs_table}). In SCAN, the mT5-base model improves the generalization accuracy over the same model architecture trained from scratch by $\sim$13\% and the larger mT5-xl model improves it by over $\sim$71\%. Similarly large improvements are observed in the COGS benchmark as well (Table~\ref{cogs_table}).

These results suggest that pretraining data do not have to come primarily from English to enable successful transfer to SCAN and COGS. However, it is still possible that the relatively small amount of English examples in mT5's training corpus might be responsible for the successful transfer to SCAN and COGS. To rule out this possibility, we also considered a model pretrained exclusively on non-English data, namely a German-to-French machine translation model (Marian NMT) trained on the OPUS-MT corpus \citep{tiedemann2020}. We observed substantial pretraining benefits even for this model (\texttt{marian\_defr} in Tables~\ref{scan_table}-\ref{cogs_table}): $\sim$11\% absolute improvement in generalization accuracy in SCAN over a baseline model with the same architecture trained from scratch and $\sim$21\% absolute improvement in generalization accuracy in COGS. This surprising result suggests that language overlap between the pretraining data and the downstream domain is not necessary for successful transfer and casts doubt on the hypothesis of \cite{furrer2020} that pretraining benefits primarily originate from an improved ability of the model to substitute similar words or phrases with each other.

To probe the limits of successful transfer further, we next asked whether pretraining with programming languages, as opposed to natural languages, would also improve generalization in downstream semantic parsing tasks. To test this, we took a recent language model called CodeT5 (denoted by \texttt{ct5\_base} in Tables~\ref{scan_table}-\ref{cogs_table}), which was pretrained predominantly on several different programming languages \citep{wang2021}. We note that the pretraining data for this model involved some amount of natural language as well, however, so the model was not pretrained exclusively with programming languages (for more details on the pretraining data and the pretraining tasks for this model, please see \cite{wang2021}). Remarkably, CodeT5 also substantially improved generalization in both SCAN and COGS (Tables~\ref{scan_table}-\ref{cogs_table}). The overall improvements were roughly equivalent in size to the improvements afforded by large-scale pretraining with natural languages. This result is consistent with an earlier observation made by \cite{papadimitriou2020} that pretraining with Java code improves downstream language modeling tasks with natural languages and it once again demonstrates the surprisingly broad transferability of the linguistic representations and knowledge learned through large-scale pretraining.

{\bf 3. Pretraining in a different domain (protein sequence modeling) mostly hurts compositional generalization:} To push the envelope even further, we next considered whether pretraining with \textit{language} corpora (natural or programming languages) was necessary for successful transfer. To test this, we used T5-base and T5-3b sized models pretrained with a protein sequence modeling task \citep{elnaggar2020}. These models are denoted by \texttt{pt5\_base} and \texttt{pt5\_xl} in Tables~\ref{scan_table}-\ref{cogs_table}. To accommodate the tokenization difference between the protein modeling and natural language modeling domains, we replaced the original protein tokenizer in \texttt{pt5\_base} and \texttt{pt5\_xl} with the corresponding T5 English tokenizer from \cite{raffel2020} and learned the token embeddings from scratch when finetuning the models on SCAN and COGS. Pretraining with the protein modeling task mostly hurt the generalization accuracy in SCAN and COGS (Tables~\ref{scan_table}-\ref{cogs_table}). For example, in SCAN, \texttt{pt5\_base} had $\sim$11\% lower absolute generalization accuracy than the corresponding T5 model trained from scratch (\texttt{t5\_base\_scr}). Similarly, in COGS, \texttt{pt5\_base} had $\sim$16\% lower generalization accuracy than \texttt{t5\_base\_scr}. These results suggest that the generalization benefits of pretraining, although broad, are not universal and that there has to be a certain kind and degree of similarity between the pretraining domain and the downstream task for successful transfer to occur.  

{\bf 4. Bigger models are harder to train from scratch, but pretraining benefits are clearer for bigger models:} In all our experiments, we consistently observed that larger models were harder to train from scratch and their generalization accuracy was lower when trained to convergence, but they also benefited more from large-scale pretraining. For example, comparing the T5-base and T5-3b models trained from scratch (\texttt{t5\_base\_scr} and \texttt{t5\_3b\_scr}), they achieve an average generalization accuracy of $\sim$43\% and $\sim$3\%, respectively, in SCAN; whereas large-scale pretraining on C4 increases these numbers to $\sim$65\% and $\sim$76\%, an absolute improvement of $\sim$23\% and $\sim$73\%, respectively (Table~\ref{scan_table}). A similar pattern is observed in COGS as well (Table~\ref{cogs_table}). These results are not entirely unexpected: bigger models are likely more prone to overfitting on relatively small-scale datasets such as SCAN and COGS, but they are much more effective at encoding the large amount of knowledge inherent in very large-scale datasets such as C4.

\section{Discussion}
Our results suggest that there is considerable generality in the transferability of pretrained representations and knowledge. We have found that even models pretrained exclusively with non-English languages or models pretrained predominantly with programming languages show substantial out-of-distribution generalization benefits in downstream semantic parsing tasks couched entirely in English (SCAN and COGS). These results add to an emerging literature documenting surprisingly broad transferability between seemingly distant pretraining and downstream tasks in both NLP and computer vision \citep{papadimitriou2020,chiang2020,lu2021,krishna2021,sinha2021,baradad2021,pondenkandath2018,maennel2020}.

However, successful transfer does not appear to be automatic or universal, potentially contrary to some recent claims \citep{lu2021}. Rather, it is likely that there are constraints on the similarity between the pretraining domain and the downstream task for successful transfer. These constraints do not have to be very stringent. For example, \cite{chiang2020} argue that an approximate match between the vocabulary sizes in the pretraining domain and the downstream task is necessary and sufficient for successful transfer. Our results are consistent with this hypothesis as we have found that a diverse range of pretrained models all with large vocabulary sizes (\texttt{t5}, \texttt{mt5}, \texttt{ct5}, \texttt{marian\_defr}) transfer successfully to the downstream semantic parsing tasks considered here, whereas pretrained protein sequence models (\texttt{pt5}) with a much smaller vocabulary size do not. 

We hope that future work will delineate these conditions for successful transfer more thoroughly. An important question to consider for future work is to what extent these conditions are content-specific. Recent work by \cite{csordas2021} suggests that a few simple content-independent modifications to transformer models, such as adjusting the scale of the embedding weights and using embeddings that are more sensitive to the relative position of tokens, can dramatically improve compositional generalization in these models when they are trained from scratch. It remains to be seen whether, or to what extent, the generalization benefits of large-scale pretraining in different tasks can potentially be explained through such relatively content-independent effects.

It seems plausible that in order to bring out the real value of pretrained knowledge and pretrained representations (as opposed to some relatively generic, content-nonspecific effects), downstream transfer tasks should be significantly more challenging than those commonly used today. In this regard, zero-shot or few-shot evaluations are probably much better than finetuning on datasets with relatively large training sets \citep{brown2020, ramesh2021}.

\bibliography{semantic_parsing}
\bibliographystyle{apalike}

\end{document}